\pdfoutput=1

\documentclass[11pt]{article}

\usepackage[]{acl}

\usepackage{times}
\usepackage{latexsym}

\usepackage[T1]{fontenc}

\usepackage[utf8]{inputenc}

\usepackage{microtype}

\usepackage{inconsolata}
\usepackage{authblk}
\usepackage{arydshln}
\usepackage{subcaption}
\usepackage{graphicx}
\usepackage{float}
\usepackage{paralist}
\usepackage{amsmath}
\usepackage{url}
\usepackage{booktabs}
\usepackage{multirow}
\usepackage{paralist}
\usepackage{enumitem}
\newcommand\sbullet[1][.5]{\mathbin{\vcenter{\hbox{\scalebox{#1}{$\bullet$}}}}}
\makeatletter
\newcommand\email[2][]%
   {\newaffiltrue\let\AB@blk@and\AB@pand
      \if\relax#1\relax\def\AB@note{\AB@thenote}\else\def\AB@note{\relax}%
        \setcounter{Maxaffil}{0}\fi
      \begingroup
        \let\protect\@unexpandable@protect
        \def\thanks{\protect\thanks}\def\footnote{\protect\footnote}%
        \@temptokena=\expandafter{\AB@authors}%
        {\def\\{\protect\\\protect\Affilfont}\xdef\AB@temp{#2}}%
         \xdef\AB@authors{\the\@temptokena\AB@las\AB@au@str
         \protect\\[\affilsep]\protect\Affilfont\AB@temp}%
         \gdef\AB@las{}\gdef\AB@au@str{}%
        {\def\\{, \ignorespaces}\xdef\AB@temp{#2}}%
        \@temptokena=\expandafter{\AB@affillist}%
        \xdef\AB@affillist{\the\@temptokena \AB@affilsep
          \AB@affilnote{}\protect\Affilfont\AB@temp}%
      \endgroup
       \let\AB@affilsep\AB@affilsepx
}
\makeatother

\title{Promptly Predicting Structures: The Return of Inference}

\author[$\spadesuit$]{\bf Maitrey Mehta}
\author[$\clubsuit$,$\diamondsuit$]{\bf Valentina Pyatkin}
\author[$\spadesuit$]{\bf Vivek Srikumar}
\affil[$\spadesuit$]{Kahlert School of Computing, University of Utah}
\affil[$\diamondsuit$]{University of Washington}
\affil[$\clubsuit$]{Allen Institute for AI}
\email{\texttt{\{maitrey,svivek\}@cs.utah.edu}, \texttt{valentinap@allenai.org}}

\begin{document}
\maketitle

\begin{abstract}
Prompt-based methods have been used extensively across NLP to build zero- and few-shot label predictors.
Many NLP tasks are naturally structured: that is, their outputs consist of \emph{multiple} labels which constrain each other. 
Annotating data for such tasks can be cumbersome.
\emph{Can the promise of the prompt-based paradigm be extended to such structured outputs?}
In this paper, we present a framework for constructing zero- and few-shot linguistic structure predictors. Our key insight is that we can use structural constraints---and combinatorial inference derived from them---to filter out inconsistent structures predicted by large language models. 
We instantiated this framework on two structured prediction tasks, and five datasets. Across all cases,
our results show that enforcing  consistency not only constructs structurally valid outputs, but also improves performance over the unconstrained variants.

\end{abstract}

\section{Introduction}
\label{sec:intro}

Structured prediction requires making multiple inter-dependent structurally constrained decisions~\cite{smith2010linguistic,nowozin2014advanced}. Many NLP tasks are naturally structured. Consider, for instance, the task of Semantic Role Labeling~(SRL) that aims to identify the semantic roles played by sentence constituents for a predicate~\cite{palmer2010semantic}. Given a sentence `\emph{Elrond gave Aragorn the sword.}' and predicate \emph{gave}, the task entails extracting its semantic arguments --- the giver~(\texttt{Agent}), the receiver~(\texttt{Recipient}), and the thing given~(\texttt{Theme}). These decisions are mutually constrained; e.g., the arguments have to be distinct phrases.

\begin{figure}
  \centering
\begin{subfigure}{7.5cm}  
 \includegraphics[width=0.8\textwidth]{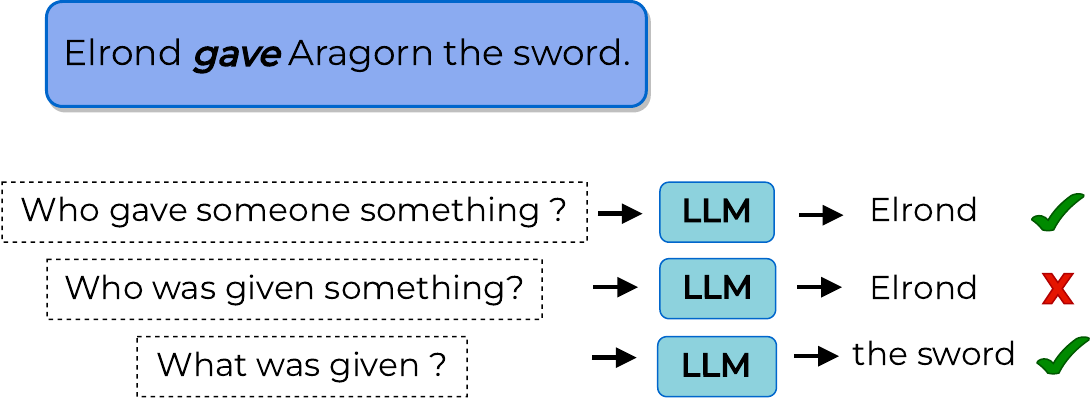}
 \caption{}
 \label{hero_image1}
\end{subfigure}
\bigskip
\begin{subfigure}{7.5cm}  
 \includegraphics[width=1\textwidth]{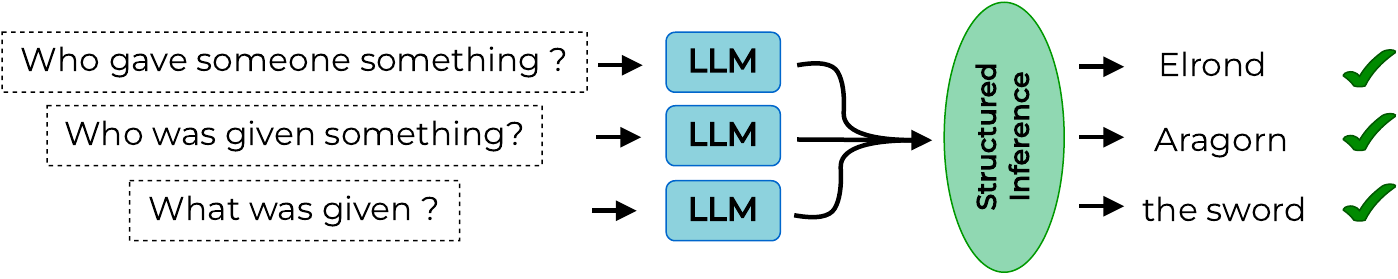}
 \caption{}
 \label{hero_image2}
\end{subfigure}
\caption{Example of Question Answer driven Semantic Role Labeling (QA-SRL) (a) without, and (b) with structured inference. Sans inference, prediction for each question may contain overlapping/repeated answers, which is prohibited per the task definition. Structured inference avoids such structurally invalid outputs.}
\label{fig1}
\end{figure}

Existing methods for structured prediction rely on large labeled datasets. However, obtaining structured labels is not straightforward, especially from non-experts. It invariably requires detailed annotation guidelines about the task, the label set, and the interactions between labels. 
Methods like the re-formulation of structured prediction tasks into a question-answering (QA) format~\cite{he-etal-2015-question,levy-etal-2017-zero,du2020event, pyatkin-etal-2020-qadiscourse, pyatkin-etal-2023-design} intend to make such tasks more amenable for annotation. However, procuring large-scale annotations still remains expensive and time-intensive; with the problem even more acute for low-resource domains. This predicament necessitates focus on zero and few-shot methods for structured prediction.

Recently, prompt-based methods~\cite[e.g.,][inter alia]{schick2021exploiting, schick-schutze-2021-just, liu2023pre} have shown immense promise, requiring few or even no labeled examples for competitive performance. Prompt-based models generate output text conditionally based on textual input called \emph{prompts}. Prompts can be crafted using a pre-defined task-specific template~\cite{raffel2020exploring}, or even just as natural language instructions~\cite{chung2022scaling} or questions~\cite{gardner2019question, pyatkin-etal-2021-asking}. 

Prompt-based methods have shown promise for text classification and text generation tasks. However, their use for structured prediction is hitherto largely unexplored. Prior work~\cite[e.g., see][]{liu2023pre} uses prompts to predict components of structures independently, but their interactions are ignored. Figure~\ref{hero_image1} illustrates a failure of that agenda. Each question pertains to a semantic role for the highlighted predicate. Since the model independently encounters each question, it runs the risk of predicting structurally inconsistent answers. 

In this paper, we propose to reintroduce inference into prompt-based methods for structured prediction problems (Figure~\ref{hero_image2}). We present a framework that allows using well-studied inference algorithms with prompt-based predictions. Inference algorithms---realized as beam search, integer linear programs~\cite[ILPs, e.g.,][]{roth-yih-2004-linear}, weighted graph optimization~\cite[e.g.,][]{tackstrom2015efficeint}, etc.---take scored candidate label sub-structures as input and optimize a global score while satisfying structural constraints. Historically, trained models generated the label scores~\cite[e.g.,][]{lafferty2001conditional,collins2002discriminative}. We argue that large language models~(LLMs), with appropriate prompts, can be used for scoring, thus eliminating the need for any training or fine-tuning. 

We instantiate this idea on two English structured prediction tasks across five datasets. We explore various facets of prompt engineering (e.g., zero- vs few-shot, instruction tuning, etc.) and their impacts on performance. We show that current models can produce structurally inconsistent outputs, and inference  helps improve not only consistency, but also overall task performance.\footnote{Code available at \url{https://github.com/utahnlp/prompts-for-structures} } 
\section{Related Work}
\paragraph{Prompts for Predictions. }
The ``pretrain, prompt, and predict'' paradigm~\cite[cf.][]{liu2023pre}
exploits the observation that LLMs yield general-purpose tools applicable to several natural language tasks. These LLMs~(e.g., T5~\cite{raffel2020exploring}, GPT-3~\cite{brownfew2020}) may be pretrained for various objectives, including 
the use of labeled data transformed into prompt-answer pairs. 
\citet{wei2021finetuned} introduced \emph{instruction tuning} that transforms natural language tasks into natural language instructions to generalize across tasks independent of their presence in training.  
With fine-tuning still possible, the prompting paradigm allows for generalization across tasks with some~(few-shot) or even no~(zero-shot) labeled samples. Several works have shown promise on zero- and few-shot text classification and generation tasks~\cite[e.g.,][inter alia]{raffel2020exploring, brownfew2020, schick-schutze-2021-just, lu-etal-2022-fantastically, touvron2023llama}. 

\paragraph{Structures \& Prompts.}
Many NLP tasks---e.g., parsing, coreference resolution, information extraction---call for predicting structured outputs such as sequences, trees or graphs. Structured prediction has a rich history in NLP~\cite{smith2010linguistic}.

Predicting structures with prompts is challenging; the output corresponds to multiple mutually-constraining decisions. 
Structures that can be flattened into label sequences 
can be cast as text generation tasks amenable to the LLM interface. \citet{blevins2022prompting} present a \emph{structured prompting} paradigm to generate labels in an auto-regressive fashion for part-of-speech tagging, named entity recognition, and sentence chunking tasks. They show that this approach works well in a few-shot setting. Such tasks are also amenable for templated slot filling in a few shot setting~\cite{cui-etal-2021-template}. \citet{liu-etal-2022-autoregressive} propose an autoregressive structured prediction framework that reformulates the problem into predicting a series of structure-building action steps.   

Structured prediction tasks can be made prompt-friendly by identifying output components and framing each one into natural instructions~\cite{he-etal-2015-question,du2020event, pyatkin-etal-2020-qadiscourse, klein-etal-2022-qasem, pyatkin-etal-2023-design} using the QA format. \citet{mekala2022zerotop} decompose semantic parsing into a series of questions. However, they do not use inference and instead, aggregate outputs sequentially. \citet{yang-etal-2022-gpt} use QA prompts to classify pairs of entity mentions as coreferent or not without aiming to form entity clusters. 
The near contemporary work of \citet{lin-etal-2023-global} presents a prompt-driven heuristic for event type classification. \citet{wang-etal-2023-code4struct} use a programming language-based prompt and generate approach for zero and few-shot event argument extraction.
Recently, \citet{le2023large} show two techniques based on QA and document infilling for zero-shot coreference resolution that rely on recent LLMs' ability to ingest entire documents as input. The concurrent work of \citet{rajaby-faghihi-kordjamshidi-2024-consistent} enforces structural consistency at inference-time, incorporating local decisions from different models using various normalizer functions.

\section{``Promptly'' Predicting Structures}
\label{section:framework}
\paragraph{Problem Statement and Notation}
Given an input $X$  (e.g., phrase, sentence, etc.), the goal of a conditional model is to characterize the distribution $P(Y|X)$ over outputs $Y$.
The outputs $Y$ in structured prediction tasks consist of a set of decisions $y_1,y_2, \cdots$. For instance, consider the SRL task from~\S\ref{sec:intro}. Given a predicate, identifying the token spans correponding to its arguments requires multiple decisions; i.e., each $y_i$ corresponds to a semantic argument. 

Since the size of the output $Y$ is variable and depends on the input $X$, the standard approach for modeling such problems calls for factorizing the distribution $P(Y|X)$ over its components $y_1, y_2, \cdots$. The factorization is a design choice; the simplest one decomposes the probability as a product over individual local decisions: $P(Y|X) = \prod_{i} P(y_i | X)$. 
With such a factorization, the prediction $Y^*$ for an input $X$ is given by 
\begin{align}
Y^* & = \max_{Y} \prod_{y_i \in Y} P(y_i \mid X) \label{eq:unconstrained-prediction}
\end{align}
These local probabilities relate to unary potentials in factor graph notation, and in our work, will correspond to logits from neural models. Importantly, the maximization problem above is trivial to solve. Each $y_i$ can be independently assigned.\footnote{Of course, the structured prediction literature studies other task-dependent factorizations. With neural models, auto-regressive factorizations are also common, where each label is conditioned on previously predicted ones. In this work, we study the simple factorization described here because it can be easily adapted to work with prompting.}

Even the simple factorization above presents two drawbacks: 
\begin{inparaenum} [i) ]
\item It requires an explicit training step using expensive-to-acquire training data to obtain the unary potentials.
\item It does not guarantee structural consistency, leading to invalid structures.
\end{inparaenum}
In this work, we ask: \emph{Can we predict valid structured outputs in the zero- to few-shot regime?}

Now, let us address these drawbacks, leading to our proposed framework.

\paragraph{Unary Potentials from Prompts.}
Prompt-based methods can help bypass the need for supervised training. For every local decision $y_i$, we construct a query $q_i$ to convert the decision into a prompt-friendly format. A pretrained model can use these queries in the standard prompt-and-predict fashion to get scored candidate answers. Formally, we write $P(Y|X,Q) = \prod_{i} P(y_i | X, q_i) $.
Here, $Q$ is a set of questions corresponding to the components of the structure. In the case of SRL, $X$ would be the input sentence and the predicate, and $Q$ would contain one question per semantic argument type for that predicate. Figure~\ref{fig1} gives an example.

\paragraph{The Return of Inference.} Merely aggregating outputs generated with prompts does not address the second drawback, i.e., the need for structurally valid outputs. Structural validity is a constraint that, when applied to the prediction problem, ensures that only valid output collections are considered as candidate predictions. 
In other words, instead of the unconstrained prediction of Eq.~\ref{eq:unconstrained-prediction}, we have 
\begin{align}
\begin{split}
    Y^*  & = \max_{Y} \prod_{y_i \in Y} P(y_i | X, q_i) \\
         & \quad\quad\quad\text{s.t.}~~ \text{Y is structurally consistent} \label{eq:constrained-prediction}
\end{split}
\end{align}
The inference problem above requires combinatorial search over valid structures.  
Inference can be realized algorithmically in various forms like beam search, an ILP~\cite{roth-yih-2004-linear} or weighted graph search~\cite[e.g.,][]{tackstrom2015efficeint, zaratiana-etal-2022-named}.

\paragraph{The Framework.} We propose the following blueprint for predicting valid structures in a zero-shot manner:  
\begin{inparaenum} [i) ]
\item Break the prediction task into its constituent sub-tasks.
\item Convert each sub-task into a prompt-friendly format.
\item Implement an inference algorithm that takes scores from the LLM and produces a valid structured output.  
\end{inparaenum}
This recipe naturally extends to a few-shot setting by prefixing in-context examples with the original prompt.  

Inference constructs outputs that satisfy constraints inherent in the definition of a task. Moreover, it can help correct errors of the zero- or few-shot models using other predictions of the same model. Without this mechanism (i.e., using the labels from the prompts directly as in Equation~\ref{eq:unconstrained-prediction}), not only will we have partially wrong outputs, they may be invalid from the perspective of the task. Finally, as we will see in the case of the coreference task, inference can help make long-range decisions about parts of an input even when the entire input is not presented to an LLM. In this sense, it can be seen as a mechanism to ``bring back the context'' without explicitly having to use a long-document model such as LongFormer~\cite{beltagy2020longformer}.

\section{Experiments}
\label{sec:experiments}

This section describes the tasks and datasets where we instantiate the framework from section~\S\ref{section:framework}.

\subsection{Semantic Role Labeling}\label{sec:srl}
Semantic Role Labeling (SRL) is the task of constructing predicate-argument structures triggered by verbs in a sentence. This task has been studied in using the PropBank~\cite{palmer-etal-2005-proposition}, FrameNet~\cite{baker-etal-1998-berkeley-framenet}, and more recently, the QA-SRL~\cite{he-etal-2015-question} datasets. 

\paragraph{Datasets.}  The QA-SRL project~\cite{he-etal-2015-question}  recasts the argument labeling task as question-answering to make it amenable for non-expert annotation. By breaking the predicate-argument structure into a collection of individual arguments, the project
curated a collection of QA pairs that represent SRL frames.   
This SRL-to-QA transformation sets it up perfectly for prompting; the sentence along with each question can serve as a prompt.

We consider two QA-SRL datasets for our experiments. 
\begin{inparaenum}[1) ]
\item \textbf{QA-SRL$_{wiki}$}~\cite{he-etal-2015-question} contains $\sim$10.8k QA pairs for 4440 predicates in 1959 sentences across splits.\footnote{\citet{he-etal-2015-question} provide datasets in newswire and Wikipedia domains. We use the latter as it is larger.}  
\item \citet{fitzgerald-etal-2018-large} present a much larger crowd-sourced dataset---\textbf{QA-SRL 2.0}---containing 265k QA pairs from 64k sentences for 133k verbal predicates annotated using the QA-SRL annotation schema.
\end{inparaenum}

\paragraph{Constraints.} The SRL definition dictates that different argument spans for a predicate do not have token overlap~\cite{palmer2010semantic}, and the QA-SRL schema mandates an answer per question. 

\begin{figure}
    \centering
    \includegraphics[width=0.49\textwidth]{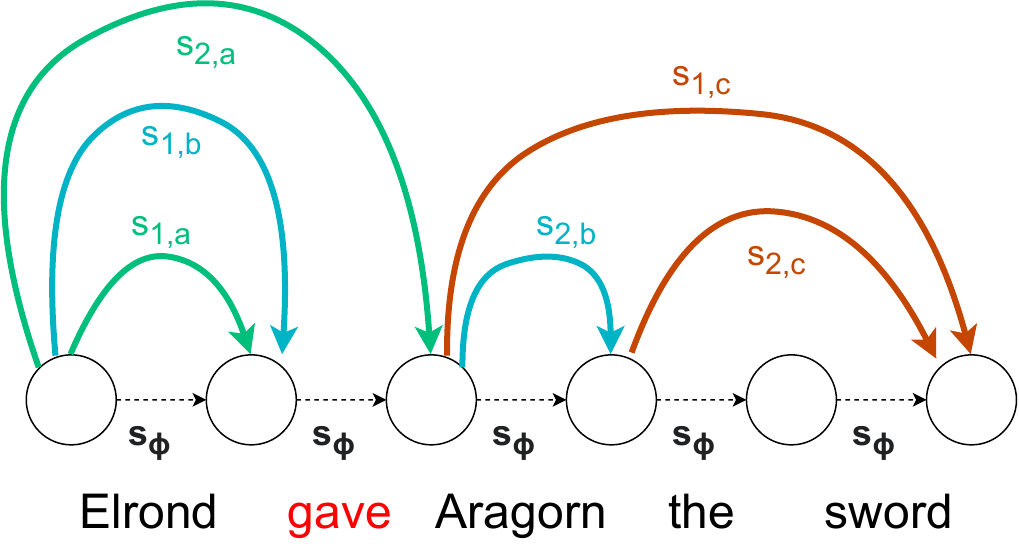}
    \caption{An example graph for the statement ``\emph{Elrond gave Aragorn the sword}''. This is a toy example with the top-2 candidate spans~($n=2$) for semantic roles=$\{a,b,c\}$. Each edge represents a candidate span. For instance, the span ranked second for scene role `a' is the phrase ``\emph{Elrond gave}'' with an edge score of $s_{2,a}$.}
    \label{fig:srl_inference}
\end{figure}

\paragraph{Prompts and Inference.} For any predicate in a sentence, we have one question $q_i$ per semantic role $i$. For every question, we prompt a generative language model to predict the best-$n$ spans (i.e., sequences) with their respective scores. We denote the score from the language model for the $r^{th}$ (where r $\leq$ n) ranked span ($t_r$) for role $i$ as $f(t_r,i)$. The scored spans from all questions for a predicate are input to an inference algorithm. 

We follow the inference algorithm of \citet{tackstrom2015efficeint}.\footnote{For brevity, we outline the inference setup here, and refer the reader to the original work for a detailed description.} Given a sentence, we construct a directed graph with one more vertex than the number of tokens. (Figure~\ref{fig:srl_inference} shows an example.)
A vertex $v_j$ is shared by two consecutive words $w_{j-1}$ and $w_j$, and denotes the boundary between them. 
Edges can be of two kinds:
\begin{inparaenum} [i) ]
    \item Null edges are added between consecutive nodes with a zero edge score.
    \item For every candidate span $t_r$ ranked $r$ for the role i, we add an edge between its representative vertices with an edge score $s_{r,i} = f(t_1,i) - f(t_r,i)$. The highest ranked span gets an edge score of zero.\footnote{A generative model tasked for extraction, like ours, cannot get indices to mark the start and end. Yet, a generated span may occur multiple times in a sentence. In such cases, edges representing all such spans are added with an identical score. } 
\end{inparaenum}

Given the graph, we can perform a shortest-path search from the leftmost to the rightmost nodes; the edges in the shortest path give us non-overlapping labels. However, doing so will generate null sequences that have a zero score. We avoid this issue by extending the algorithm to return the top $K$ shortest paths using Yen's $K$-shortest path algorithm~\cite{yen1971finding}. Subsequently, the highest scoring path among the $K$ paths which assigns exactly one span per semantic role is the optimal structure that satisfies all task constraints. 
For the example in Figure \ref{fig:srl_inference}, our inference algorithm would return `\emph{Elrond}', `\emph{Aragorn}', and `\emph{the sword}' as the arguments for roles a, b and c respectively.\footnote{See Appendix \ref{appendix:srl_worked_out} for a worked-out example.}

\paragraph{Evaluation. } We evaluate at both the question-level and the structure-level. At each level, we use exact and head match metrics~\cite[p.~49]{palmer2010semantic}. At the question-level, predicted spans are compared with gold spans. For the \textbf{Exact$_{q}$} accuracy metric, answers are considered correct if the spans exactly match. Similarly, answers are considered correct if the phrasal heads of the spans match to compute the \textbf{Head$_{q}$} accuracy metric. A structure is considered accurate if predicted spans for all constituent roles match. As at the question-level, a match can be determined in an exact (\textbf{Exact$_{s}$}), or head (\textbf{Head$_{s}$}) fashion. Additionally, we also evaluate inconsistency percent~($\rho$) by measuring how often constraints are violated. We compare every pair of outputs(i.e., argument spans) in a predicted structure. A violation occurs when two output spans overlap.

\subsection{Coreference Resolution}
\label{subsec: exp_coref}
Coreference resolution is the task of grouping mention expressions (or simply, mentions) in a document into clusters that refer to the same entity. For example, consider the sentence ``\emph{Al requested Bob to give him the pen.}'', with given mentions: $\{$``Al'', ``Bob'', ``him'',  ``the pen''$\}$. Here,  ``Al'' and ``him'' refer to the same entity, i.e., Al, while the other mentions are singletons. 

\paragraph{Datasets.} We use three datasets for this task: 
\begin{inparaenum} [1) ]
    \item The entity coreference annotations within the EventCorefBank+ dataset~\cite[ECB+,][] {cybulska-vossen-2014-using}.\footnote{We use the processed files provided by \citet{yang-etal-2022-gpt}} 
    \item The CoNLL 2012 OntoNotes 5.0 dataset with the v12 English data splits~\cite{pradhan-etal-2012-conll}, which contains entity and event coreference annotation across several domains like magazine articles, broadcast conversations, among others.
    \item The GENIA Coreference Corpus~\cite{su_et_al:DagSemProc.08131.4}, which helps showcase the applicability of this framework in a low-resource domain like biomedicine where annotation is especially difficult. 
     The GENIA corpus is a collection of coreference annotations for 2000 paper abstracts published in the biomedical domain. We define our own train/dev/test splits for the corpus since pre-defined splits were unavailable. We will release these splits for future use. For OntoNotes and GENIA, we do not consider nested entities.
\end{inparaenum}

\paragraph{Constraints.} The coreference resolution task can be broken into a series of binary decisions: for every mention pair in a document we can ask whether the mentions are co-referrant or not. Entity clusters require transitive closure. That is, given three mentions $m_i$, $m_j$ and $m_k$, if a predicate $C(x,y)$ denotes $x$ and $y$ refer to the same entity, then:
\begin{equation}\small
    \forall {i,j,k}, C(m_i,m_j) \land C(m_j,m_k) \implies C(m_i,m_k)
\label{eq:transitivity_cosnt}
\end{equation}

\paragraph{Prompt and Inference.} For every document, we collect mention pairs in it.\footnote{For Ontonotes, we considered all mention pairs in a window size of three sentences due to a large number of mentions.} We transform the coreference decision about mentions $m_i$ and $m_j$ into a Yes/No question: ``\emph{Does $m_i$ refer to $m_j$?}''. A `Yes' establishes a coreference link between the mentions. Questions are appended with the relevant context and given to a language model. Link scores~($s_{i,j}$) for a mention-pair $i$ and $j$  are given by $s_{i,j} = f_{yes}(i,j) - f_{no}(i,j) $ where $f_{yes}$ and $f_{no}$ are the scores for generating the sequences "yes" and "no" given a question and context. Using the link scores, we employ \emph{All-link} inference from \citet{chang-etal-2011-inference} to solve the following integer program over binary decisions $y_{i,j}$:
\begin{align*}\small
    \max_y & \sum_{i,j} y_{i,j}s_{i,j} \\ 
        \text{such that}\quad  & y_{i,k} \geq  y_{i,j} + y_{j,k} - 1,  ~~~\forall i,j,k \\
         &  y_{i,j} \in \{0,1\}, ~~~\forall i,j.
\end{align*}
The constraint ensures that the transitivity condition from \eqref{eq:transitivity_cosnt} is satisfied.

\paragraph{Evaluation.} We use two metrics: 
\begin{inparaenum} [i)]
    \item \textbf{F1} that measures performance of the binary decisions at the question level
    \item \textbf{CoNLL score}~\cite{pradhan-etal-2014-scoring} that averages three cluster evaluation metrics: MUC , B$^3$, and CEAF$_{e}$. 
\end{inparaenum}
Additionally, we measure the inconsistency percent~($\rho$): the number of times constraint \eqref{eq:transitivity_cosnt} is violated divided by the number of times the antecedent is true. This is the same as the `conditional violation' defined by \citet{li-etal-2019-logic}, and measures the fraction of times a third link between three mentions does not exist, when the other two links between them exist. 
\section{Results}
This section presents the impact of inference on zero-shot performance and consistency. We use the 3 billion variants from the T5~\cite{raffel2020exploring} family of encoder-decoder models for our experiments.  We discuss our implementation briefly here; Appendix~\ref{appendix:exp_details} has details like prompt templates. 

\subsection{Semantic Role Labeling}
We consider the T5-3B model\footnote{We also considered UnifiedQAv2~\cite{khashabi2022unifiedqa} and Macaw~\cite{Tafjord2021Macaw}. T5 performed the best.} for our experiments. We also present results on Flan T5-XL~\cite{wei2021finetuned}, also a 3B parameter model, to check if instruction tuning helps with zero-shot performance. 

Instruction-tuned models like Flan T5 allow verbalizing constraints within the prompt structure. We also test out an iterative prompting method~\cite{wang-etal-2022-iteratively}, where questions pertaining to a predicate in a sentence are presented in sequential order to the model with verbalized constraints. In every prompt, we include the QA pair from every previous question for the predicate already prompted. This prompting strategy is denoted by the superscript \texttt{itr}.  See Appendix~\ref{subsec: itr-prompt-template} for an example.\footnote{Note that verbalized constraints do not guarantee a valid output structure. However, exposure to preceding predictions may help reduce inconsistencies nonetheless.} Further, we can use inference in conjunction with this iterative prompting strategy.

Tables \ref{tab:srl_results_wiki} and \ref{tab:srl_results_qasrl2} show the results  for QA-SRL$_{wiki}$ and QA-SRL 2.0 respectively. We see that the constrained models produce near-consistent gains over their unconstrained counterparts. Flan-T5 does not outperform the T5 model, but iterative prompting with inference significantly improves its performance. More importantly, without inference, the models produce highly inconsistent predictions with a lowest inconsistency percent of 34\%! In addition, we also provide results on GPT-4 with the constraints verbalized as system instruction.\footnote{GPT-4 results are meant to contextualize our results with a current SOTA system. The datasets we considered may have been utilized to train GPT-4. GPT-4 also has multiple orders of magnitude more parameters than the models we consider.} For GPT-4, all the questions are provided at once. Refer Appendix~\ref{subsec: gpt4-template} for details.

As an illustrative example, we present a case where constrained inference corrects overlapping predicted arguments. For the sentence, 
\emph{``Keats' long and expensive medical training with Hammond and at Guy's Hospital led his family to assume he would pursue a lifelong career in medicine, assuring financial security \ldots''}, the model was asked:
\begin{inparaenum} [i) ]
\item What would assure something?
\item What would something assure?
\end{inparaenum}
The unconstrained model generated ``financial security'' on both counts. Post-inference, the answer to the first question is changed to ``lifelong career in medicine'', resulting in the correct structure.

\begin{table}[ht]
\centering
\begin{tabular}{l|rr|r} \hline
 \textbf{System}          &  \textbf{Head$_q$}($\uparrow$) & \textbf{Head$_s$}($\uparrow$) & $\rho$ ($\downarrow$) \\ \hline
         T5-3B             &  \textbf{63.84}  & 36.17 & 34.30\\
         ~~~ + constraints & 62.11  & \textbf{38.75} & 0 \\
         Flan-T5-XL        & 36.76  & 11.20 & 87.16 \\
         ~~~ + constraints & 43.34  & 18.48 & 0 \\ 
         Flan-T5-XL$^{itr}$        & 59.47  & 33.60 & 49.79 \\
         ~~~ + constraints         & 61.11  & 36.17 & 0 \\ \hline
         GPT-4        & 80.46  & 62.93 & 5.61 \\\hline
\end{tabular}
\caption{Semantic Role Labeling performance and consistency metrics on the QA-SRL$_{wiki}$ dataset for unconstrained vs constrained systems. A fine-tuned constrained T5 model gets a Head$_s$ of 58.12. All values in \%. Detailed metrics in Table~\ref{tab:srl_results_wiki_det} in Appendix~\ref{appendix:srl_detailed_results}.}
\label{tab:srl_results_wiki}
\end{table}

 \begin{table}
\centering
\begin{tabular}{l|rr|r} \hline
 \textbf{System}          &  \textbf{Head$_q$}($\uparrow$) & \textbf{Head$_s$}($\uparrow$) & $\rho$ ($\downarrow$) \\ \hline                               
    T5-3B             &  68.57  & 46.35 & 39.06\\
     ~~~ + constraints & 68.87  & 52.31 & 0 \\
    Flan-T5-XL        &  61.92  & 40.37  & 58.19\\
     ~~~ + constraints & 66.02  & 48.68  & 0 \\ 
    Flan-T5-XL$^{itr}$        &   68.84	& 49.29  & 42.11 \\
         ~~~ + constraints         & \textbf{71.97}  & \textbf{55.41} & 0 \\ \hline
         GPT-4        & 86.78  & 76.89 & 7.18 \\\hline
\end{tabular}
\caption{Semantic Role Labeling performance and consistency metrics on the QA-SRL 2.0 dataset for unconstrained vs constrained systems. A fine-tuned constrained T5 model gets a Head$_s$ of 69.00. All values in \%. Detailed metrics in Table~\ref{tab:srl_results_qasrl2_det} in Appendix~\ref{appendix:srl_detailed_results}. }
\label{tab:srl_results_qasrl2}
\end{table}

\subsection{Coreference Resolution}
\label{subsec:coref_results}
The ability to specify multiple choices makes the Macaw-3B~\cite{Tafjord2021Macaw} model an ideal choice for the Yes/No choice QA task.\footnote{We also considered this task in an NLI format with the T5-3B model. However, the QA format was a clear winner.} As before, we show results using the instruction-tuned Flan T5-XL model as well. 
For this set of experiments, we only prompt the model with the sentence(s) containing the mention pair as context, and not the whole document.\footnote{Note: We do not need a large context window. Inference can create consistent clusters by looking at smaller contexts.} We show more analysis with varying context styles in Appendix~\ref{appendix:context_styles}.

\begin{table}
\centering
\begin{tabular}{l|rr|r} \hline
 \textbf{System} & \textbf{F1}($\uparrow$) & \textbf{CoNLL}($\uparrow$) & \textbf{$\rho$}($\downarrow$) \\ \hline 
    All-Yes              & 13.32 &  41.51 & 0    \\ \hline
    All-No           & 45.84 &  45.06 & 0    \\ \hline
    Macaw-3B            & 46.26 &  N/A   & 48.56 \\
    ~~~ + R2L           & 42.20 &  53.55 & 0     \\
    ~~~ + All-Link      & 48.57 &  57.76 & 0     \\
    Flan-T5-XL          & 61.58 &  N/A   & 80.78  \\
    ~~~ + R2L           & 58.38 &  62.79 & 0  \\
    ~~~ + All-Link      & \textbf{66.06} &  \textbf{65.09} & 0  \\ \hline
\end{tabular}
\caption{Coreference resolution performance and consistency metrics for unconstrained vs constrained systems for the ECB+ dataset. A fine-tuned constrained Macaw model gets an F1 and CoNLL score of 85.69 and 85.14 respectively. All values in \%.}
\label{tab:coref_results_ecbp}
\end{table}

In addition to the constrained inference proposed in \S\ref{subsec: exp_coref}, we consider a simple right-to-left cluster assignment heuristic (\textbf{R2L}) proposed by \citet{soon-etal-2001-machine}.  This heuristic assigns a candidate mention to the cluster of the closest mention where a link was predicted by the unconstrained model. We also present two rudimentary baselines: 
\begin{inparaenum}[i) ]
    \item \textbf{All-Yes} baseline where all questions are answered ``Yes'' and all mentions belong to a single cluster, and
    \item \textbf{All-No} baseline where all questions are answered ``No'' and every mention is its own cluster.  
\end{inparaenum}
These baselines are relevant since both generate valid structures. 

The unconstrained and the constrained models can only be compared using their F1 scores and the inconsistencies.\footnote{The official CoNLL scorer requires valid structures rendering base system evaluation impossible w/o inference.} Tables~\ref{tab:coref_results_ecbp}, ~\ref{tab:coref_results_ontonotes} and ~\ref{tab:coref_results_genia} show results for the ECB+, OntoNotes and GENIA datasets  respectively. We refer the reader to Table~\ref{tab:coref_results} in Appendix~\ref{appendix: coref_zero_shot} for a detailed breakdown of the CoNLL score. 
We observe gains in F1 scores for constrained systems over their unconstrained counterparts. In all cases, our models outperform the All-Yes and All-No baselines. We also see that the inconsistency is substantially higher for the unconstrained models, especially with Flan-T5-XL.

\begin{table}
\centering
\begin{tabular}{l|rr|r} \hline
\textbf{System} & \textbf{F1}($\uparrow$) & \textbf{CoNLL}($\uparrow$) & \textbf{$\rho$}($\downarrow$) \\ \hline 
    All-Yes           & 18.96 &  38.61 & 0  \\ \hline
    All-No            & 43.38 &  39.42 & 0  \\ \hline
    Macaw-3B          & 48.67 &  N/A      & 63.88   \\
    ~~~ + R2L         & 49.48 & \textbf{55.30} & 0  \\
    ~~~ + All-Link & \textbf{52.15} &  55.14 & 0  \\
    Flan-T5-XL        & 48.93 &  N/A      & 86.75   \\
    ~~~ + R2L         & 51.88 & 50.37 & 0  \\
    ~~~ + All-Link & 51.33 & 48.52 & 0  \\ \hline
\end{tabular}
\caption{Coreference resolution performance and consistency metrics for unconstrained vs constrained systems for the OntoNotes dataset. A fine-tuned constrained Macaw model gets an F1 and CoNLL score of 96.17 and 95.63 respectively. All values in \%.}
\label{tab:coref_results_ontonotes}
\end{table}

\begin{table}
\centering
\begin{tabular}{l|rr|r} \hline
\textbf{System} & \textbf{F1}($\uparrow$) & \textbf{CoNLL}($\uparrow$) & \textbf{$\rho$}($\downarrow$) \\ \hline 
    All-Yes           &  9.00 & 37.70 & 0  \\ \hline
    All-No            & 47.40 & 31.19 & 0  \\ \hline
    Macaw-3B          & 46.63 & N/A      & 51.18  \\
    ~~~ + R2L         & 35.62 & 47.25 & 0  \\
    ~~~ + All-Link    & 46.75 & 52.56 & 0  \\
    Flan-T5-XL        & 56.53 & N/A      & 82.64  \\
    ~~~ + R2L         & 53.35 & 54.11 & 0  \\
    ~~~ + All-Link & \textbf{65.36} &  \textbf{57.47} & 0  \\ \hline
\end{tabular}
\caption{Coreference resolution performance and consistency metrics for unconstrained vs constrained systems for the GENIA dataset. A fine-tuned constrained Macaw model gets an F1 and CoNLL score of 91.58 and 90.01 respectively. All values in \%.}
\label{tab:coref_results_genia}
\end{table}

\begin{figure}
    \centering
    \includegraphics[width=0.5\textwidth]{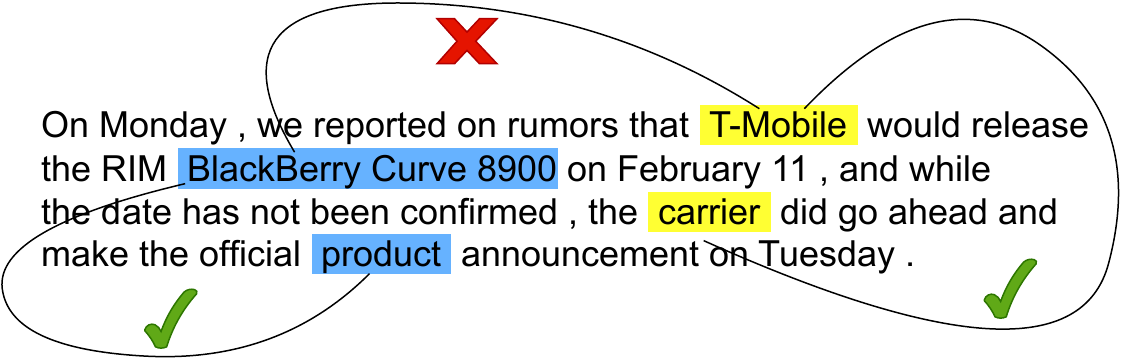}
    \caption{An example from ECB+ dataset where inference helps correct inconsistency of predictions. An incorrect link is predicted between `\emph{Blackberry Curve 8900}' and `\emph{T-Mobile}' by the unconstrained model, which is removed post-inference. Irrelevant mentions are hidden for clarity.}
    \label{fig:success_story_cref}
\end{figure}

 Figure~\ref{fig:success_story_cref} presents an instance where inference corrects flawed links. Here, `\emph{T-mobile}' and `\emph{carrier}' belong to one entity cluster, while, `\emph{Blackberry Curve 8900}' and `\emph{product}' belong to another. Our unconstrained model incorrectly predicts a link between `\emph{T-mobile}' and `\emph{Blackberry Curve 8900}'.  But transitivity requires all other edges to be present, which the model does not predict. Our inference algorithm removes the erroneous edge and predicts consistent clusters.

Iterative prompting for coreference resolution is challenging for two reasons: 
\begin{inparaenum}[1)]
    \item As the component questions establish links between two mentions and the transitivity closure applies to three mentions, verbalizing the transitivity constraint within the prompt can be challenging.
    \item Coreference structures are much larger than the SRL structures, hence, demanding a large memory requirement. 
\end{inparaenum} 

\section{Analysis Experiments}
\label{sec:analysis}
Next, we show how performance and inconsistency vary with model size and in-context examples.

\begin{figure}[ht]
    \centering
    \includegraphics[width=0.5\textwidth]{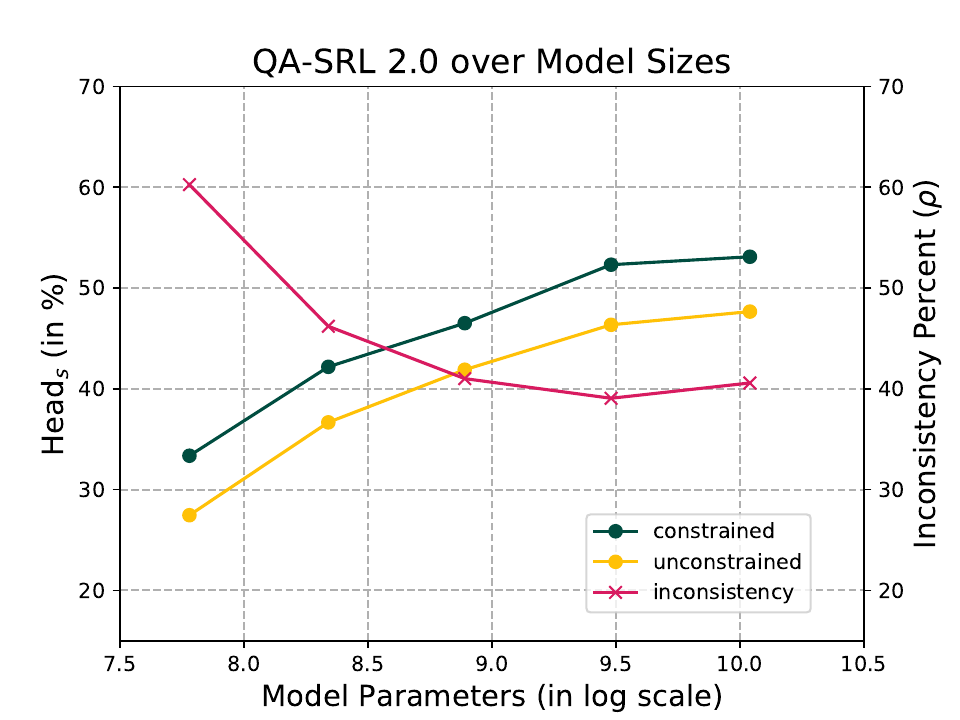}
    \caption{Head$_s$ performance and inconsistency percent over model sizes for the QA-SRL 2.0 dataset.  The circle-marked (-$\sbullet[.75]$-) plots map to the left axis, and the cross-marked (-x-) plot to the right axis. Inconsistency is shown for unconstrained models since constrained models are always consistent (i.e., $\rho$=0).}
    \label{fig:qasrl2_model_size}
\end{figure}

\begin{figure}[ht]
    \centering
    \includegraphics[width=0.5\textwidth]{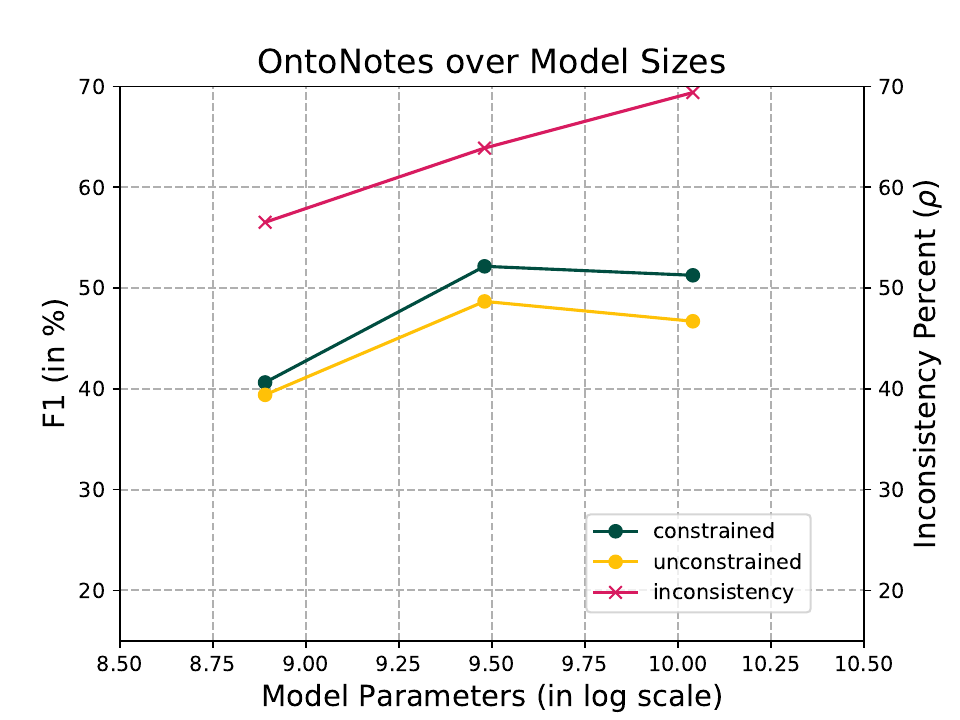}
    \caption{F1 performance and inconsistency percent over model sizes for the OntoNotes dataset. The visual encodings follow the same convention as in Figure~\ref{fig:qasrl2_model_size}. }
    \label{fig:onto_model_size}
\end{figure}

\begin{figure}[ht]
    \centering
    \includegraphics[width=0.5\textwidth]{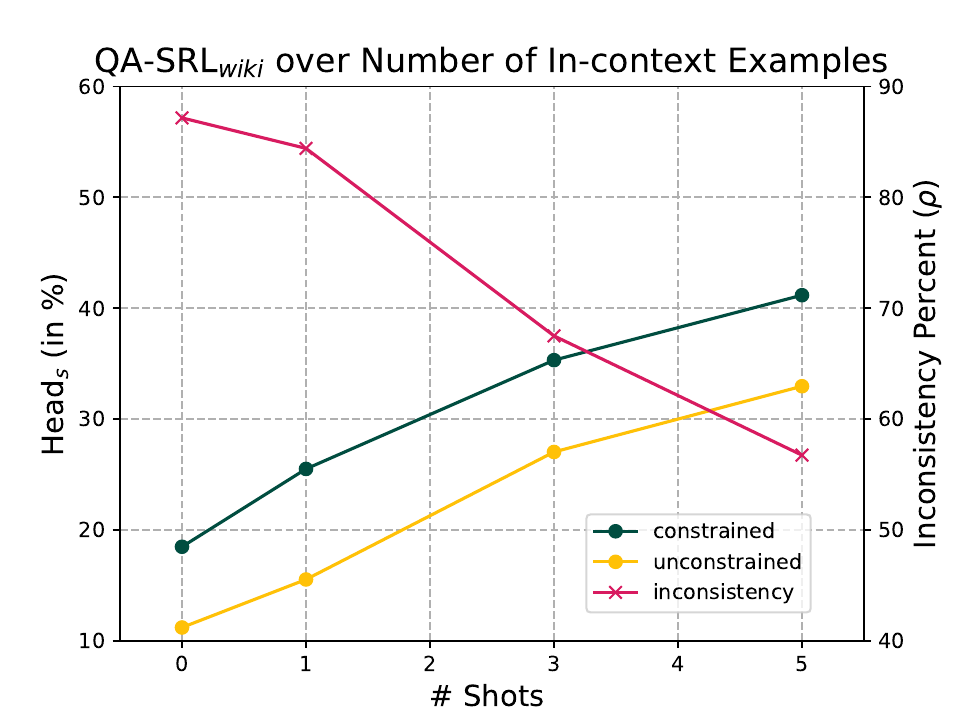}
    \caption{Head$_s$ performance and inconsistency percent over in-context examples (shots) for the QA-SRL$_{wiki}$ dataset. The visual encodings follow the same convention as in Figure~\ref{fig:qasrl2_model_size}.}
    \label{fig:wiki_few_shot}
\end{figure}

\begin{figure}[ht]
    \centering
    \includegraphics[width=0.5\textwidth]{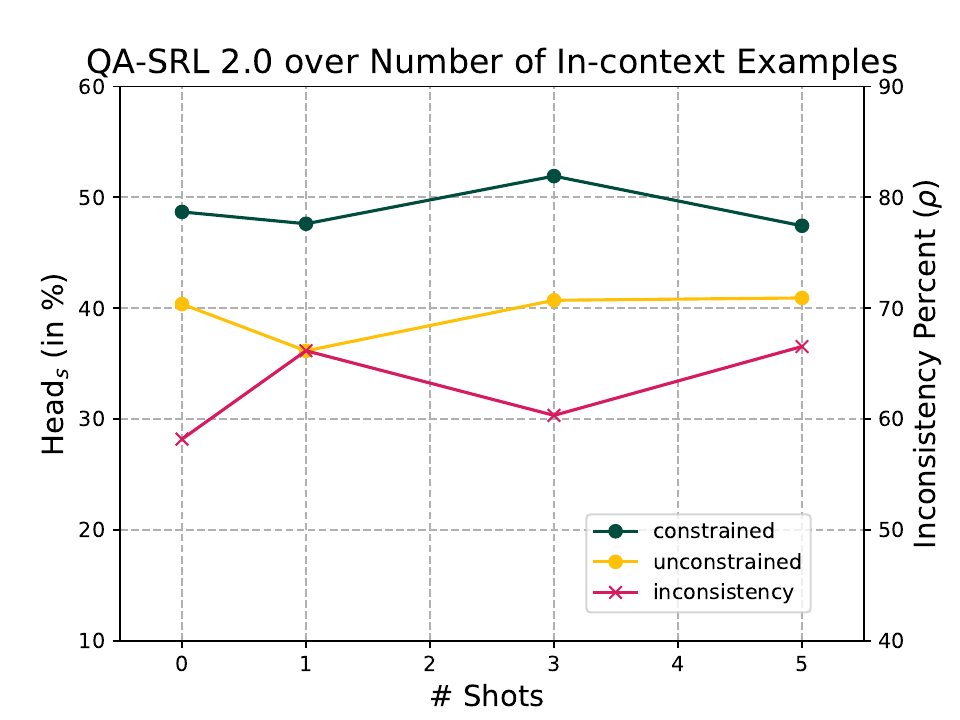}
    \caption{Head$_s$ performance and inconsistency percent over in-context examples (shots) for the QA-SRL 2.0 dataset. The visual encodings follow the same convention as in Figure~\ref{fig:qasrl2_model_size}.}
    \label{fig:qasrl2_few_shot}
\end{figure}

\subsection{Model Size}
We study the performance and consistency of models over varying sizes. 
For SRL, we compare  the small, base, large, 3 billion, and 11 billion T5 models. Figure~\ref{fig:qasrl2_model_size} shows  the Head$_s$ metric over T5 model sizes for QA-SRL 2.0, along with the inconsistency percent.  
For coreference, we compare the large, 3 billion and 11 billion variants of the Macaw model. We consider the All-Link inference for the constrained models. We report the F1 score and the inconsistency percent on the OntoNotes dataset in Figure~\ref{fig:onto_model_size}. 
The inconsistency is shown only for the unconstrained models; the constrained ones are guaranteed to be consistent.

Across the datasets, we see that performance almost always increases and constraints provide steady gains. In QA-SRL 2.0, inconsistency reduces with increasing model size. The trend reverses for OntoNotes. Interestingly, with QA-SRL 2.0, a constrained model of a certain size often yields comparable, if not better, results as an unconstrained model of the immediately bigger model, showing that constraints might give performance gains equivalent to using a bigger model. Appendix~\ref{appendix:model_size} shows results for the other datasets.

\subsection{Few-shot Prompting}

So far, we have examined our framework in the zero-shot setting. \emph{Do the conclusions hold for few-shot settings as well?} 
To answer this question, we use the Flan-T5-XL model which is known to yield promising results with even a few in-context examples.\footnote{An interesting question: What counts as a shot? All questions in a structure or just one? We consider the latter.} We conduct few-shot experiments for the SRL task on the QA-SRL$_{wiki}$ (Figure~\ref{fig:wiki_few_shot}) and QA-SRL 2.0 (Figure~\ref{fig:qasrl2_few_shot}). We observe a clear increase in performance with increasing shots for QA-SRL$_{wiki}$. Recall that Flan T5, in a zero-shot setting for this dataset, underperforms compared to T5. Hence, the gains are not surprising: only the 5-shot constrained system beats the zero-shot constrained T5 model (38.75 Head$_s$). The model also becomes more consistent with more examples. For QA-SRL 2.0, the performance and the consistency largely remain invariant with number of shots. Constrained systems produce more accurate structures than their unconstrained counterparts.

\section{Conclusions}
Prompt-based methods have impressed when deployed in zero- and few-shot manner. Structured prediction problems can use them by breaking the structure down into smaller components and querying for local decisions. This aspect has been explored by some in the literature. However, these works leave out a pivotal part of such structured prediction tasks---the structure. In this work, we proposed a new framework to predict structurally consistent outputs by following local predictions with an inference step. In our experiments, our framework not only guarantees consistency but also provides consistent performance gains.
\section*{Limitations}
Our focus of this work is to predict structures given that a structured prediciton task can be broken down into components and questions can be generated for those components. While this relies heavily on advances in question generation, we do not focus on this line of research and consider that questions are given to us. A separate line of work exists precisely in this domain of question generation for structured prediction tasks \cite{he-etal-2015-question, levy-etal-2017-zero, du2020event, wu-etal-2020-corefqa, pyatkin-etal-2020-qadiscourse, klein-etal-2020-qanom, pyatkin-etal-2021-asking}. We emphasize that our contribution is more general as to how we can marry the idea of inference with prompting.

A problem that can affect any generative model is the label bias problem where the model prefers a certain label over certain other label irrespective of the inputs. This problem can cause performance imbalances and models need to be calibrated to alleviate this phenomenon. Recently, a few works~\cite{zhao2021calibrate, holtzman-etal-2021-surface,chen2022close, han2022prototypical, fei2023mitigating} study this phenomenon for zero and few shot prompting models for an array of text classification tasks. We briefly studied this for our coreference experiments and found that Macaw was more well-calibrated for the binary task as compared to T5. We leave a detailed calibration study for future work.

\section*{Acknowledgements}
We thank the members at the Utah NLP lab and AI2 for their insights, especially, Ashim Gupta for his feedback on the iterative prompting experiments. We thank the anonymous reviewers for their feedback. This material is based upon work supported in part by NSF under grants \#2007398, \#2217154, \#1822877 and \#1801446. The support and resources from the Center for High Performance Computing at the University of Utah are gratefully acknowledged. Valentina is supported by an Eric and Wendy Schmidt Postdoctoral Award.
\bibliography{main}

\appendix
\section{Detailed SRL Results}
\label{appendix:srl_detailed_results}
Detailed results for the QA-SRL$_{wiki}$ and QA-SRL 2.0 are given in Tables \ref{tab:srl_results_wiki_det} and \ref{tab:srl_results_qasrl2_det} respectively.

\begin{table*}
\centering
\begin{tabular}{l|rrrr|r} \hline
 \textbf{System}          & \textbf{Exact$_q$}($\uparrow$) & \textbf{Exact$_s$}($\uparrow$)  & \textbf{Head$_q$}($\uparrow$) & \textbf{Head$_s$}($\uparrow$) & $\rho$ ($\downarrow$) \\ \hline
         T5-3B             & 38.80  & 12.65  & \textbf{63.84}  & 36.17 & 34.30\\
         ~~~ + constraints & 40.30  & 14.78  & 62.11  & \textbf{38.75} & 0 \\
         Flan-T5-XL        & 22.72  &  3.47  & 36.76  & 11.20 & 87.16 \\
         ~~~ + constraints & 33.12  &  9.52  & 43.34  & 18.48 & 0 \\ 
         Flan-T5-XL$^{itr}$   & 37.57 &  12.32    & 59.47  & 33.60 & 49.79 \\
         ~~~ + constraints    & \textbf{43.80} & \textbf{17.25}    & 61.11  & 36.17 & 0 \\ \hline
         GPT-4             & 62.84  &  37.07  & 80.46  & 62.93 & 5.61 \\ \hline
\end{tabular}
\caption{Semantic Role Labeling performance and consistency metrics on the QA-SRL$_{wiki}$ dataset for unconstrained vs constrained systems. A fine-tuned constrained T5 model gets a Head$_s$ of 58.12. All values in \%. }
\label{tab:srl_results_wiki_det}
\end{table*}

 \begin{table*}
\centering
\begin{tabular}{l|rrrr|r} \hline
 \textbf{System}          & \textbf{Exact$_q$}($\uparrow$) & \textbf{Exact$_s$}($\uparrow$)  & \textbf{Head$_q$}($\uparrow$) & \textbf{Head$_s$}($\uparrow$) & $\rho$ ($\downarrow$) \\ \hline                               
    T5-3B             & 54.12  & 30.49  & 68.57  & 46.35 & 39.06\\
     ~~~ + constraints & 56.78  & 36.96  & 68.87  & 52.31 & 0 \\
    Flan-T5-XL        & 48.92  & 26.53  & 61.92  & 40.37  & 58.19\\
     ~~~ + constraints & 57.06  & 37.23  & 66.02  & 48.68  & 0 \\ 
     Flan-T5-XL$^{itr}$    &  55.11 & 32.86    &   68.84	& 49.29  & 42.11 \\
         ~~~ + constraints     & \textbf{61.34} & \textbf{41.17}    & \textbf{71.97}  & \textbf{55.41} & 0 \\ \hline
    GPT-4             & 77.98  &  63.00  & 86.78  & 76.89 & 7.18 \\ \hline
\end{tabular}
\caption{Semantic Role Labeling performance and consistency metrics on the QA-SRL 2.0 dataset for unconstrained vs constrained systems. A fine-tuned constrained T5 model gets a Head$_s$ of 69.00. All values in \%. }
\label{tab:srl_results_qasrl2_det}
\end{table*}

\section{Additional Coreference Results}
\label{appendix: coref_results}

\subsection{Zero Shot Coreference Results}
\label{appendix: coref_zero_shot}
The complete zero shot results for Macaw-3B and Flan T5-XL models are given in Table \ref{tab:coref_results}. One might notice that the Flan model is highly inconsistent as compared to the Macaw model. Note that in each case, the divisor is different since it is the number of times the antecedent is activated in equation \ref{eq:transitivity_cosnt}. Examining the predictions, we saw that the value of the divisor for the Macaw models is roughly four times that of the Flan models. This implies that Flan-T5 tends to predict `No' more often.

\begin{table*}
\centering
\begin{tabular}{ll|rrrrr|r} \hline
\textbf{Dataset} & \textbf{System} & \textbf{F1}($\uparrow$) & \textbf{MUC}($\uparrow$) & \textbf{B$^3$}($\uparrow$) & \textbf{CEAF$_e$}($\uparrow$) & \textbf{CoNLL}($\uparrow$) & \textbf{$\rho$}($\downarrow$) \\ \hline 
                    & Single            & 13.32 & \textbf{63.93} & 41.67 & 18.94 & 41.51 & 0    \\ \cdashline{2-8}
                    & Singleton         & 45.84 &  0.00 & 74.13 & 61.05 & 45.06 & 0    \\ \cdashline{2-8}
                    & Macaw-3B          & 46.26 &  N/A    &  N/A     &   N/A    & N/A      & 48.56 \\
  \multirow{2}{*}{ECB+}   & ~~~ + R2L   & 42.20 & 52.71 & 61.00 & 46.93 & 53.55 & 0     \\
                    & ~~~ + All-Link & 48.57 & 56.72 & 65.59 & 50.98 & 57.76 & 0     \\
                    & Flan-T5-XL        & 61.58 & N/A      & N/A      & N/A      & N/A      & 80.78  \\
                    & ~~~ + R2L         & 58.38 & 53.03 & 73.27 & 62.08 & 62.79 & 0  \\
                    & ~~~ + All-Link & \textbf{66.06} & 51.59 & \textbf{77.07} & \textbf{66.60} & \textbf{65.09} & 0  \\ \hline
                    & Single            & 18.96 & \textbf{70.23} & 30.89 & 14.71 & 38.61 & 0  \\ \cdashline{2-8}
                    & Singleton         & 43.38 &  0.00 & 66.43 & 51.84 & 39.42 & 0  \\ \cdashline{2-8}
                    & Macaw-3B          & 48.67 & N/A      & N/A      & N/A      & N/A      & 63.88   \\
  \multirow{2}{*}{Ontonotes}& ~~~ + R2L & 49.48 & 51.11 & 66.04 & 48.76 & \textbf{55.30} & 0  \\
                    & ~~~ + All-Link & \textbf{52.15} & 47.67 & 67.02 & 50.72 & 55.14 & 0  \\
                    & Flan-T5-XL        & 48.93 & N/A     & N/A      & N/A      & N/A      & 86.75   \\
                    & ~~~ + R2L         & 51.88 & 31.12 & \textbf{67.51} & 52.47 & 50.37 & 0  \\
                    & ~~~ + All-Link & 51.33 & 24.93 & 67.08 & \textbf{53.56} & 48.52 & 0  \\ \hline
                   & Single             &  9.00 & \textbf{77.41} & 26.81 &  8.89 & 37.70 & 0  \\ \cdashline{2-8}
                    & Singleton         & 47.40 &  0.00 & 57.10 & 36.47 & 31.19 & 0  \\ \cdashline{2-8}
                    & Macaw-3B          & 46.63 & N/A     & N/A      & N/A      & N/A      & 51.18  \\
  \multirow{2}{*}{GENIA} & ~~~ + R2L    & 35.62 & 66.00 & 45.11 & 30.65 & 47.25 & 0  \\
                    & ~~~ + All-Link & 46.75 & 62.40 & 56.19 & 39.08 & 52.56 & 0  \\
                    & Flan-T5-XL        & 56.53 &  N/A     & N/A      & N/A      & N/A      & 82.64  \\
                    & ~~~ + R2L         & 53.35 & 59.62 & 58.30 & 44.41 & 54.11 & 0  \\
                    & ~~~ + All-Link & \textbf{65.36} & 53.48 & \textbf{66.90} & \textbf{52.04} & \textbf{57.47} & 0  \\ \hline
\end{tabular}
\caption{Coreference resolution performance and consistency metrics for unconstrained vs constrained systems. All values in \%.}
\label{tab:coref_results}
\end{table*}

\subsection{Context Styles for Coreference Resolution}
\label{appendix:context_styles}
As mentioned in \S\ref{subsec:coref_results}, for any given question pertaining to a mention-pair, we consider only the sentences where the constituent mentions occur. One can argue that surrounding context can play an important role in deciding whether or not two mentions are co-referent. Furthermore, intermediate context between the sentences can be pivotal in establishing otherwise loosely connected sentences appearing a few sentences apart. This begs the question about how much, if any, does providing the entire document help.\footnote{For Ontonotes and GENIA, the document size went beyond memory which is why we do not test along this dimension.} On a different note, mentions can be highlighted with asterisks(as used for WSC in \citet{raffel2020exploring}) to explicitly state where in the context they occur. This approach can help deal with pronominal mentions which are bound to repeat. We present results when the entire document~(\textbf{Full}) acts as context, when mentions are highlighted~(\textbf{Hlght}), and a combination of both~(\textbf{Full + Hlght}) in comparison to the context style where only the relevant sentences are considered~(\textbf{Rel}).  We provide detailed results for performance and consistency over differing context styles in Table \ref{tab:coref_context_style}. We see that results are comparable across all context styles.

\begin{table*}
\centering
\begin{tabular}{ll|rrrrr|r} \hline
\textbf{Dataset} & \textbf{System} & \textbf{F1}($\uparrow$) & \textbf{MUC}($\uparrow$) & \textbf{B$^3$}($\uparrow$) & \textbf{CEAF$_e$}($\uparrow$) & \textbf{CoNLL}($\uparrow$) & \textbf{$\rho$}($\downarrow$) \\ \hline 
                    & Rel          & 46.26 &  N/A    &  N/A     &   N/A    & N/A      & 48.56 \\
  \multirow{2}{*}{ECB+}   & ~~~ + R2L   & 42.20 & 52.71 & 61.00 & 46.93 & 53.55 & 0     \\
                    & ~~~ + All-Link & \textbf{48.57} & 56.72 & \textbf{65.59} & \textbf{50.98} & \textbf{57.76} & 0     \\
                    & Hlght             & 46.03 & N/A      & N/A      & N/A      & N/A      & 46.68  \\
                    & ~~~ + R2L         & 42.49 & 51.19 & 60.76 & 47.17 & 53.04 & 0  \\
                    & ~~~ + All-Link & 47.84 & 57.92 & 64.80 & 49.59 & 57.44 & 0  \\ 
                    & Full              & 44.47 &  N/A    &  N/A     &   N/A    & N/A      & 31.15 \\
                    & ~~~ + R2L         & 38.46 & 57.51 & 60.25 & 43.97 & 53.91 & 0     \\
                    & ~~~ + All-Link & 45.59 & 59.28 & 64.69 & 48.93 & 57.63 & 0     \\
                    & Full+Hlght        & 43.54 & N/A      & N/A      & N/A      & N/A      & 31.02  \\
                    & ~~~ + R2L         & 37.38 & 59.17 & 59.44 & 43.15 & 53.92 & 0  \\
                    & ~~~ + All-Link & 44.66 & \textbf{60.34} & 63.80 & 48.56 & 57.57 & 0  \\ \hline
                    & Rel               & 48.67 & N/A      & N/A      & N/A      & N/A      & 63.88   \\
  \multirow{2}{*}{Ontonotes}& ~~~ + R2L & 49.48 & 51.11 & 66.04 & 48.76 & 55.30 & 0  \\
                    & ~~~ + All-Link & 52.15 & 47.67 & 67.02 & 50.72 & 55.14 & 0  \\
                    & Hlght             & 49.22 & N/A     & N/A      & N/A      & N/A      & 64.93   \\
                    & ~~~ + R2L         & 49.40 & \textbf{51.16} & 65.88 & 48.75 & 55.26 & 0  \\
                    & ~~~ + All-Link & \textbf{52.43} & 48.20 & \textbf{67.13} & \textbf{50.90} & \textbf{55.41} & 0  \\ \hline
                    & Rel               & 46.63 & N/A     & N/A      & N/A      & N/A      & 51.18  \\
  \multirow{2}{*}{GENIA} & ~~~ + R2L    & 35.62 & 66.00 & 45.11 & 30.65 & 47.25 & 0  \\
                    & ~~~ + All-Link & \textbf{46.75} & 62.40 & \textbf{56.19} & 39.08 & 52.56 & 0  \\
                    & Hlght             & 46.55 &  N/A     & N/A      & N/A      & N/A      & 50.62  \\
                    & ~~~ + R2L         & 34.94 & \textbf{66.29} & 45.30 & 30.81 & 47.47 & 0  \\
                    & ~~~ + All-Link & 46.66 & 62.67 & 55.92 & \textbf{39.22} & \textbf{52.60} & 0  \\ \hline
\end{tabular}
\caption{Coreference resolution performance and consistency metrics over different context styles for unconstrained vs constrained systems. The first row of every triplet denotes the context style considered. All values in \%. All results are on the Macaw-3B model.}
\label{tab:coref_context_style}
\end{table*}

\section{Additional Model Size Experiments}
\label{appendix:model_size}

\begin{figure}
    \centering
    \includegraphics[width=0.5\textwidth]{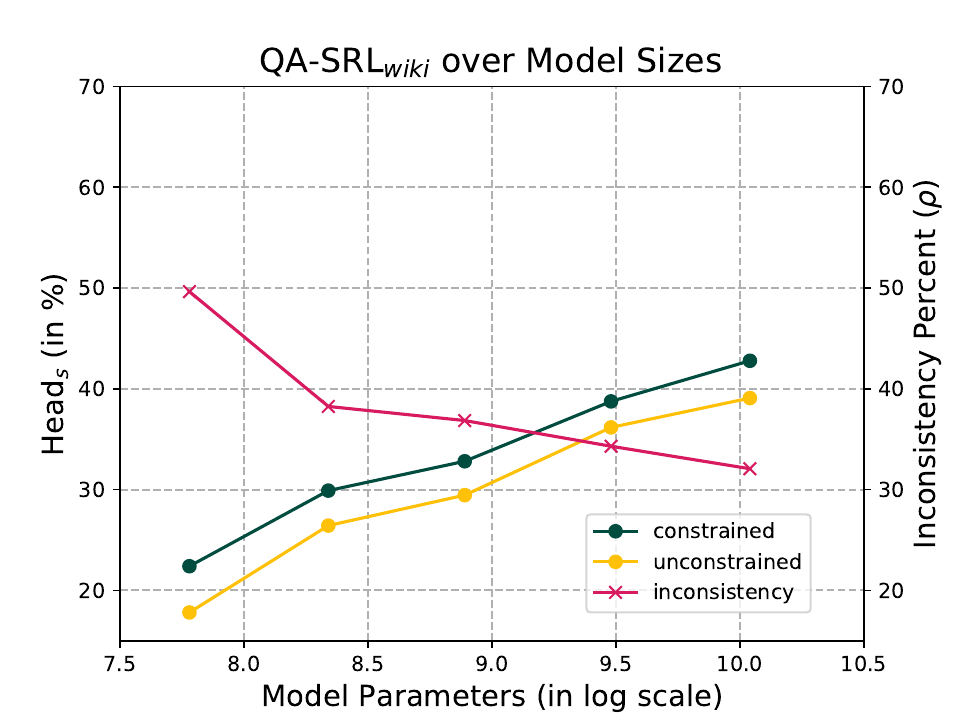}
    \caption{Head$_s$ performance and inconsistency percent over model sizes for the QA-SRL$_{wiki}$ dataset. Refer  circle-marked (-$\sbullet[.75]$-) plots to the left axis, and the cross-marked (-x-) plot to the right axis. Inconsistency is computed for unconstrained models since constrained models are trivially consistent (i.e., $\rho$=0).}
    \label{fig:wiki_model_size}
\end{figure}

The analysis over model sizes for QA-SRL$_{wiki}$ is given in Figure \ref{fig:wiki_model_size}. We see similar analysis as observed with QA-SRL 2.0 dataset. Performance improves with increasing model size while inconsistency reduces. Furthermore, constraints provide steady gains over their unconstrained counterparts.

\begin{figure}
    \centering
    \includegraphics[width=0.5\textwidth]{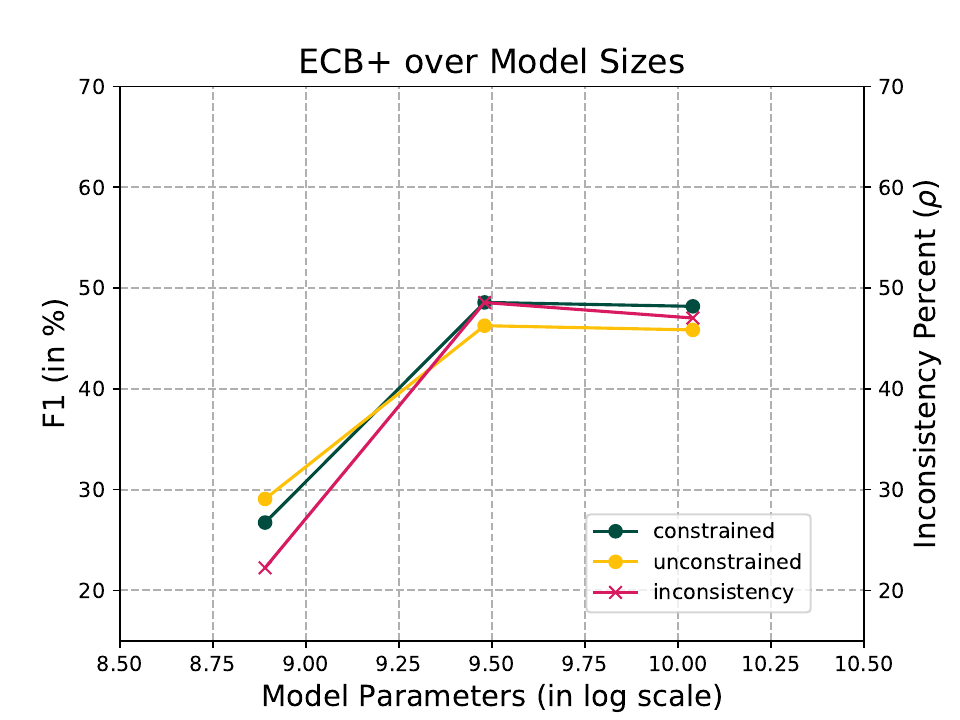}
    \caption{F1 performance and inconsistency percent over model sizes for the ECB+ dataset. Refer  circle-marked (-$\sbullet[.75]$-) plots to the left axis, and the cross-marked (-x-) plot to the right axis. Inconsistency is computed for unconstrained models since constrained models are trivially consistent (i.e., $\rho$=0).}
    \label{fig:ecbp_model_size}
\end{figure}

\begin{figure}
    \centering
    \includegraphics[width=0.5\textwidth]{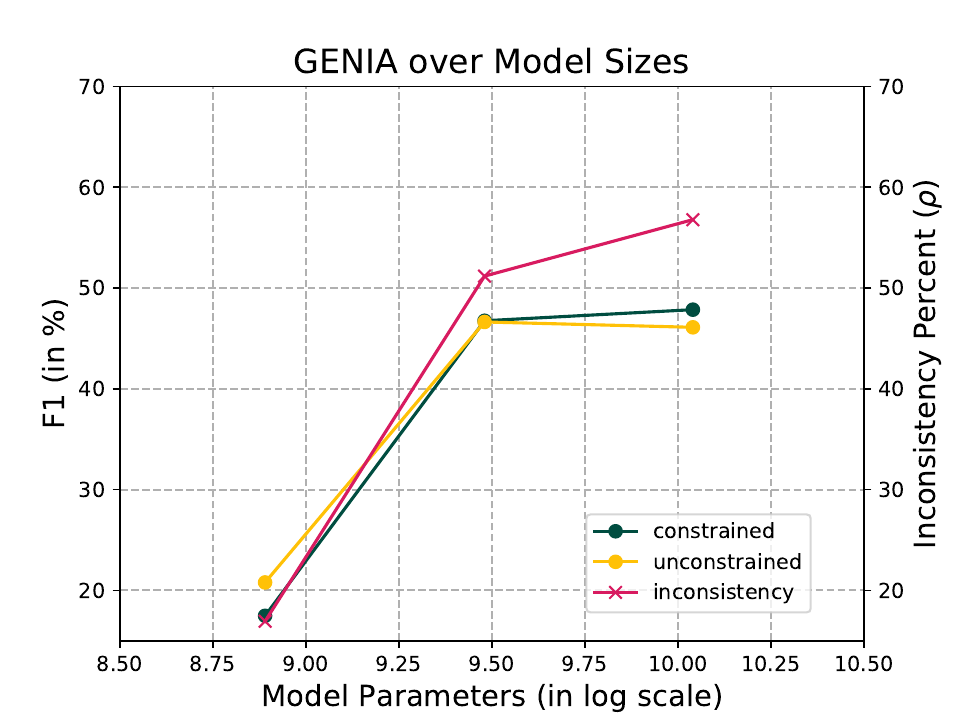}
    \caption{F1 performance and inconsistency percent over model sizes for the GENIA dataset. Refer  circle-marked (-$\sbullet[.75]$-) plots to the left axis, and the cross-marked (-x-) plot to the right axis. Inconsistency is computed for unconstrained models since constrained models are trivially consistent (i.e., $\rho$=0).}
    \label{fig:genia_model_size}
\end{figure}

Figure \ref{fig:ecbp_model_size} shows performance and inconsistency of the Macaw model over sizes for the ECB+ coreference dataset. Figure \ref{fig:genia_model_size} shows the same for the GENIA dataset. In both the cases, we see that performance drastically improves from the large to the 3 billion parameter variant. Performance gains from 3 billion to 11 billion are minimal. However, we see that the inconsistency increases with increasing size. We also see that for these datasets, for the large variant, constraints do not help the model. However, note that F1 is a question-level metric and a drop might be a necessary to guarantee a consistent structure.  

\section{Experimental and Prompt Details}
\label{appendix:exp_details}
We use HuggingFace's \texttt{transformers} library~\cite{wolf-etal-2020-transformers} for all our models. All model generations were done with a random seed of 2121.  We use the `\texttt{generate}' method in the HuggingFace~\cite{wolf-etal-2020-transformers} library to return the score of a generated sequence. These scores can be accessed by the `\texttt{sequences\_scores}' output attribute. We consider the top 20 spans for each question. For all few shot settings, we average results across shots considered from three seeds: 42, 20 and 1984.

We explain the prompt templates and the parameters/assumptions we chose in the subsequent sections.

\subsection{Semantic Role Labeling.} 
We use the questions and sentences provided by the respective datasets. Say, the question and the sentence context were denoted by <ques> and <context> respectively. For the T5 experiments, we follow the QA prompt format as specified by \citet{raffel2020exploring} which is: ``\texttt{question: <ques> context: <context>}''. For Flan-T5, we use the following prompt design: ``\texttt{<context> \textbackslash n In the above sentence, <ques>}''. These are the inputs to the respective models. For generation, we considered a beam size of 20. We considered 20 shortest paths to be returned from Yen's K-shortest path algorithm. Generally, a higher K might get better results but we found 20 to be optimal for the perfomance-time tradeoff.

\subsubsection{Iterative Prompting Template}
\label{subsec: itr-prompt-template}
\begin{figure}
    \centering
    \includegraphics[width=0.5\textwidth]{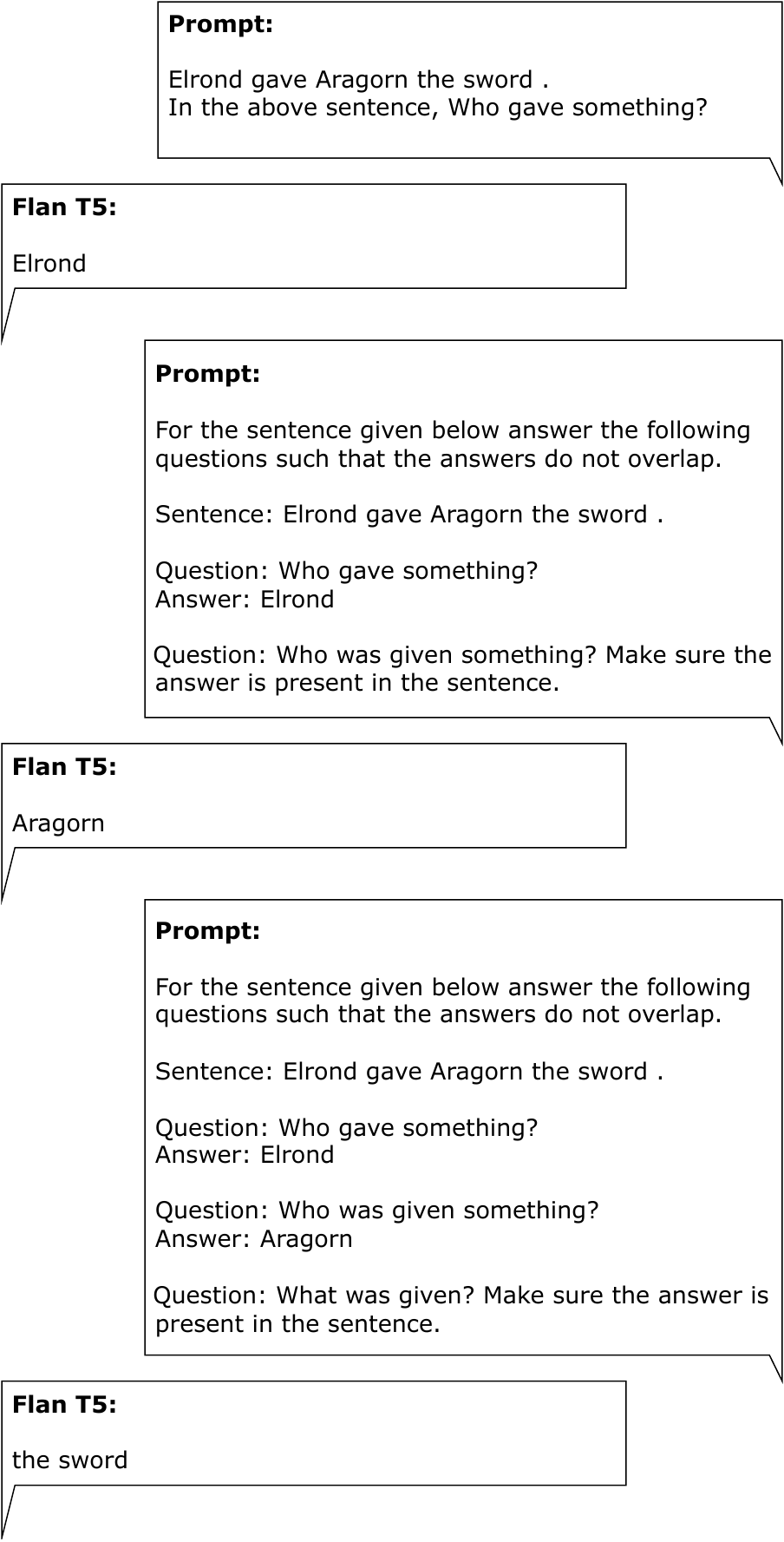}
    \caption{An example of the iterative prompting template. Note that Flan-T5 is not a stateful model.}
    \label{fig:itr_template}
\end{figure}

In the iterative prompting template, we prompt the model one question at a time such that the answer for every preceding prompt is included in subsequent prompts. Assume a predicate in a sentence to have $n$ questions: \{q$_1$,q$_2$,\ldots,q$_n$\}. In the first step, we prompt the model for q$_1$ using the prompt template mentioned in the previous sub-section. Next, we follow with the prompt the model for q$_2$ such that the question and the generated answer for q$_1$ are included in the prompt. Consequently, we use a different prompt template that contains this pair and an instruction that asks the answers to not overlap. Subsequently, the prompt for q$_3$ will contain the question-answer pair q$_1$ and q$_2$, and so on. An example is shown in Figure \ref{fig:itr_template}. We use the order provided in the QA-SRL datasets for our experiments.

\subsubsection{GPT-4 Prompt Template}
\label{subsec: gpt4-template}

\begin{figure}
    \centering
    \includegraphics[width=0.5\textwidth]{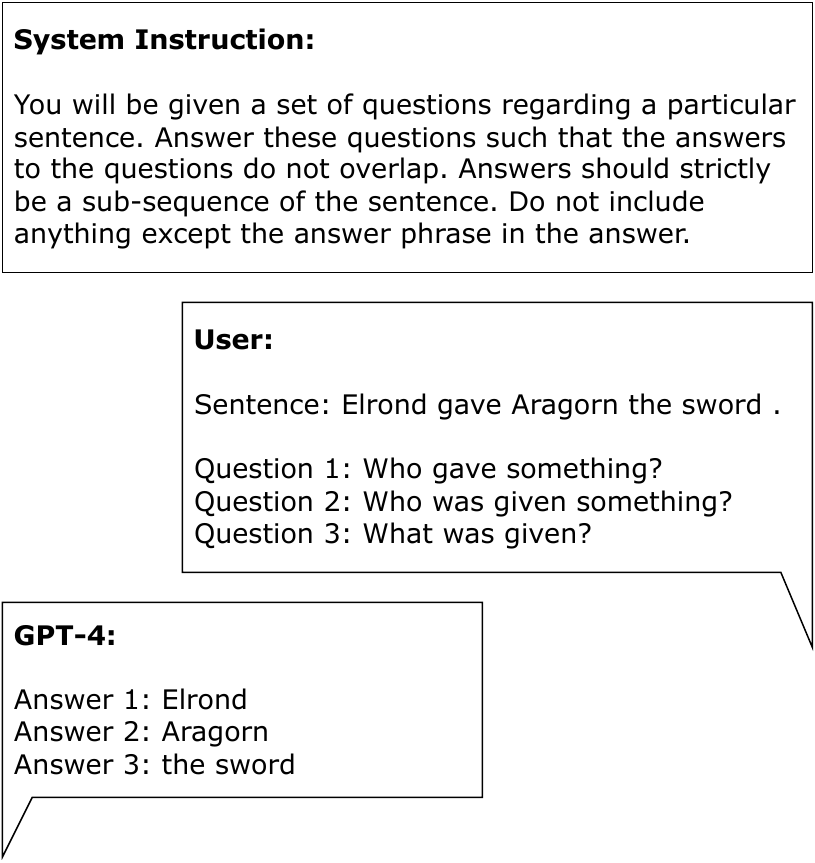}
    \caption{An example with the template used for GPT-4 experiments.}
    \label{fig:gpt4_template}
\end{figure}

We used OpenAI's API and the \texttt{ChatCompletion} end-point to obtain the GPT-4 results. We provide the following system instruction: ``\emph{You will be given a set of questions regarding a particular sentence. Answer these questions such that the answers to the questions do not overlap. Answers should strictly be a sub-sequence of the sentence. Do not include anything except the answer phrase in the answer}''. A user prompt consists of a sentence followed by a set of questions pertaining to a certain predicate in the sentence. As a result, each predicate consists of a request to the endpoint. An example request and answer are shown in Figure \ref{fig:gpt4_template}. A majority of the GPT-4 generations adhere to the answer template shown in the figure. However, for predicates with a single question, answers did not always contain answer headers ( e.g., ``\texttt{Answer 1: <answer>}''). These were handled accordingly. In some rare cases, ``\texttt{None of the above choices}'' was generated and such cases were considered as having no answers. The experiment across the two datasets cost a total of \$96.31.

\subsection{Coreference Resolution.} 
We are given the context (<context>) and a pair of mentions(<m1> and <m2>) for the input. For the Macaw models, the input prompt follows this format: \\``\texttt{\$answer\$ ; \$mcoptions\$=(A) Yes (B) No  ; <context> Does <m1> refer to <m2>?}'' \\
For the  Flan models, we follow the format: 
\\``\texttt{<context> \textbackslash n In the above passage, does <m1> refer to <m2>? Yes or No?}''\\
We restrict generations to \{``Yes'',``No''\} for Flan experiments and to \{``\$answer\$ = Yes'', ``\$answer\$ = No''\} for the Macaw experiments. Integer Linear Programs were implemented and solved using the Gurobi solver~\cite{gurobi}.

\section{Inference Overhead}
Since the inference algorithm follows the generation step, a time overhead is expected. Table \ref{tab:time_tradeoff} shows the generation time and its corresponding inference time for the datasets we consider. Crucially, while the generation step is run on the GPU, our inference implementation is run on the CPU.

\begin{table}
\centering
\begin{tabular}{l|rr} \hline
 \textbf{Dataset}          &  \textbf{Gen. step}& \textbf{Inf. Step}\\ \hline
         QA-SRL$_{wiki}$            &  48.49  &  2.18 \\
         QA-SRL 2.0                 & 945.22  & 32.68  \\
         ECB+                       &  81.87  &  0.08  \\
         Ontonotes                  &1547.03  & 62.99  \\ 
         GENIA                      & 859.88  &116.33   \\\hline
\end{tabular}
\caption{Time taken (in minutes) by the generation~(Gen.) and the inference~(Inf.) steps for the datasets. The generation step for QA-SRL$_{wiki}$, QA-SRL 2.0, and Ontonotes were executed on an Nvidia A100 (40 GB) GPU. The generation step for ECB+ and GENIA were executed on an Nvidia A40 (48 GB) GPU. The benchmarking is performed for the T5-3B and Macaw-3B models for the SRL and coreference resolution tasks respectively.}
\label{tab:time_tradeoff}
\end{table}
\section{SRL Inference Worked-out Example}
\label{appendix:srl_worked_out}

\begin{table*} [ht]
\centering
\begin{tabular}{cccc}
\toprule
\multirow{2}{*}{\textbf{Rank}} & \multicolumn{3}{c}{\textbf{Role}}    \\ \cmidrule(l){2-4} 
                               & \textbf{a} & \textbf{b} & \textbf{c} \\ \midrule
1        &     ``\textit{Elrond}'' (2)        &    ``\textit{Aragorn}'' (5)     &    ``\textit{Aragorn the sword}'' (5)       \\
2        &     ``\textit{Elrond gave}'' (1)   &    ``\textit{Elrond}'' (3)      &     ``\textit{the sword}'' (4)        \\ \hline
\end{tabular}
\caption{Example spans and scores~(in brackets) of candidate spans for each role. The scores correspond to the $f(t_r,i)$ values discussed in \S\ref{sec:srl} where is $t_r$ is the r$^{th}$ ranked span for role $i$.}
\label{tab:toy_vals}
\end{table*}

\begin{figure*}[ht]
    \centering
    \includegraphics[width=0.75\textwidth]{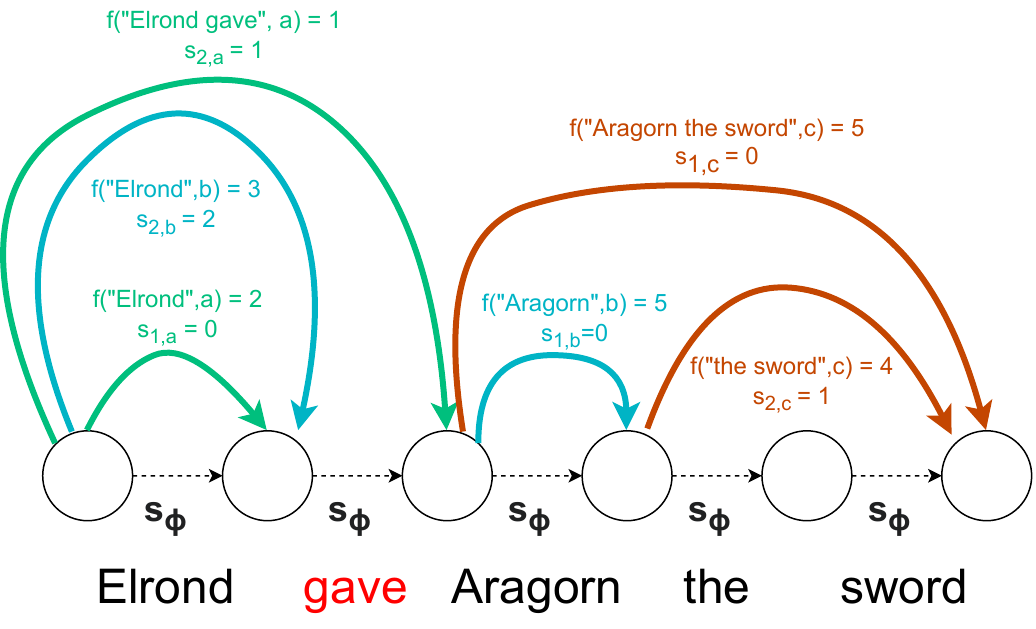}
    \caption{An example of the SRL inference graph with values. Every edge shown in the figure contains the scores returned by the scoring function (in our case, the generative language model) $f(t_r,i)$ where is $t_r$ is the r$^{th}$ ranked span for role $i$. As mentioned in \S\ref{sec:srl}, edge score for a span $t_r$ is then calculated as $s_{r,i} = f(t_1,i) - f(t_r,i)$ where $f(t_1,i)$ is the score (as returned by the scoring function) for the top ranked span for role $i$. Note that we use these scores as the edge weights and, subsequently, as the scores used in the inference algorithm.}
    \label{fig:worked-out-srl-ex}
\end{figure*}

\begin{table} [ht]
    \centering
    \begin{tabular}{l|l|c}
   \textbf{Rank} &     \textbf{Path} & \textbf{Path length}  \\ \hline
   1 & $\{s_{1,a}, s_{1,b}\}$      &     0                   \\
   2 & $\{s_{1,a}, s_{1,c}\}$      &     0                   \\
   3 & $\{s_{1,a}\}$      &     0                   \\
   4 & $\{s_{1,b}\}$      &     0                  \\
   5 & $\{s_{1,c}\}$      &     0                   \\
   6 & $\phi$  &     0                   \\
   \textbf{7} & \pmb{$\{s_{1,a}\rightarrow s_{1,b}\rightarrow s_{2,c}\}$}        &     \textbf{1}\\
   8 & $\{s_{2,a}\rightarrow s_{1,c}\}$        &     1\\
   \multicolumn{3}{c}{\ldots}\\\hline
    \end{tabular}
    \caption{K-shortest paths returned by the}
    \label{tab:paths}
\end{table}

We will consider the example mentioned in Figure \ref{fig:srl_inference} with toy values to illustrate the inference process. In this toy setup, we consider three roles (\texttt{a},\texttt{b}, and \texttt{c})\footnote{You may assume these to be the \texttt{Agent}, \texttt{Recipient} and \texttt{Theme} roles as mentioned in \S\ref{sec:intro}} and assume the top-2 candidate spans per role with their scores as returned a hypothetical generative model. 

The spans and scores for this toy example are given in Table \ref{tab:toy_vals}. Given these spans and scores. we can construct the directed graph as shown in Figure \ref{fig:worked-out-srl-ex}. We calculate the edge scores $s_{r,i} = f(t_1,i) - f(t_r,i)$ where $f(t_1,i)$ is the score (as returned by the scoring function) for the top ranked span for role $i$. For the sake of this illustration, we shall use $s_{r,i}$ as the span notation as well. The $s_{\phi}$ value is set to zero. The first step of our inference algorithm aims to find the $K$ shortest paths from the leftmost to the rightmost node. Observe how overlapping spans can never be on the same path. As a results, the top ranked candidates for roles $b$ (span $s_{1,b}$) and $c$ (span $s_{1,c}$) can never be in the same structure.

$K$ shortest paths are returned in increasing order of total path length where path length for a path S is $\sum_{s_{r,i}\in S}s_{r,i}$ as shown in Table \ref{tab:paths}. As a result, one might notice that partial structures such as paths $\{s_{1,a}\rightarrow s_{1,c}\}$ or $\{s_{1,a} \rightarrow s_{1,b}\}$ might have the shortest path length. However, the QA-SRL framework necessitates one answer per question(or, in other words, role). A post-processing step follows the shortest path step where, from the $K$ shortest paths returned, the shortest path containing exactly one span for each role is chosen. This step also ensures that paths where multiple spans for the same role are present (e.g., $\{s_{2,b}\rightarrow s_{1,b}\rightarrow s_{2,c}\}$), are also avoided. In the example shown in Figure \ref{fig:worked-out-srl-ex}, $\{s_{1,a}\rightarrow s_{1,b}\rightarrow s_{2,c}\}$ gives the most optimal path with a path length 1 and satisfies all the constraints. This yields to ``\textit{Elrond}'', ``\textit{Aragorn}'' and ``\textit{the sword}'' as arguments for roles \textit{a}, \textit{b} and \textit{c} respectively. Intuitively, a sufficiently large $K$ (in our case, 20) leads to a higher chance of obtaining a complete and consistent structure. In cases where no complete structures can be formed from any of the $K$ shortest paths, partial~(yet, consistent) structures are predicted.\footnote{Partial structures can be considered as having "null" answers for some roles. For evaluation purposes, we treat them as incorrect answers.}

\end{document}